\documentclass[runningheads]{llncs}
\usepackage{graphicx}
\usepackage{times}
\usepackage{epsfig}
\usepackage{graphicx}
\usepackage{amsmath}
\usepackage{amssymb}
\usepackage{lineno}
\usepackage{color}
\usepackage{subfigure}
\usepackage{multirow}
\usepackage{algorithm}
\usepackage{algorithmic}
\usepackage{epsfig}
\usepackage{caption}
\usepackage{bm}
\usepackage{url}

\newcommand{\wh}{\textcolor[rgb]{0,0,0}}
\newcommand{\cdn}{\textcolor[rgb]{0,0,0}}
\newcommand{\whh}{\textcolor[rgb]{0,0,0}}
\urldef{\mailsa}\path|lixiang651@gmail.com,majch7@mail2.sysu.edu.cn|
\urldef{\mailsb}\path|liweih3@mail2.sysu.edu.cn,wszheng@ieee.org|
\begin{document}

\title{Towards Photo-Realistic Visible Watermark Removal \\ with Conditional Generative Adversarial Networks}

\author{Xiang Li\inst{1} \and Chan Lu\inst{2}\and Danni Cheng\inst{3} \and Wei-Hong Li\inst{4} \protect\\ Mei Cao \inst{1} \and Bo Liu\inst{5}  
	\and Jiechao Ma\inst{1}\thanks{Corresponding author} \and Wei-Shi Zheng\inst{1}}


\institute{ 
   $^1$Sun Yat-sen University, Guangdong, China\\
   $^2$Shanghai University of Finance and Economics, Shanghai, China\\
	$^3$Shanghai Jiao Tong University, Shanghai, China\\
	$^4$The University of Edinburgh, Edinburgh, United Kingdom\\
   $^5$Zhejiang University, Zhejiang, China\\
	\mailsa\\}

\maketitle              

\begin{abstract}

\wh{Visible watermark plays an important role in image copyright protection and the robustness of a visible watermark to an attack is shown to be essential. To evaluate and improve the effectiveness of watermark, watermark removal attracts increasing attention and becomes a hot research topic. Current methods cast the watermark removal as an image-to-image translation problem where the encode-decode architectures with pixel-wise loss are adopted to transfer the transparent watermarked pixels into unmarked pixels. However, when a number of realistic images are presented, the watermarks are more likely to be unknown and diverse (i.e., the watermarks might be opaque or semi-transparent; the category and pattern of watermarks are unknown). When applying existing methods to the real-world scenarios, they mostly can not satisfactorily reconstruct the hidden information obscured under the complex and various watermarks (i.e., the residual watermark traces remain and the reconstructed images lack reality). To address this difficulty, in this paper, we present a new watermark processing framework using the conditional generative adversarial networks (cGANs) for visible watermark removal in the real-world application. The proposed method \whh{enables} the watermark removal solution \whh{to be more} closed to the photo-realistic reconstruction using a patch-based discriminator conditioned on the watermarked images, which is adversarially trained to differentiate the difference between the recovered images and original watermark-free images. Extensive experimental results on a large-scale visible watermark dataset demonstrate the effectiveness of the proposed method and clearly indicate that our proposed approach can produce more photo-realistic and convincing results compared with the state-of-the-art methods.}


\end{abstract}

\section{Introduction}

\wh{Nowadays, people tend to post photos and videos on the Internet for sharing and preserving memories of events and so on. To protect the copyright of \whh{photos and videos}, the visible watermark is commonly used. Typically, those watermarks are opaque or semi-transparent images containing names or logos, overlaying on the original images. Despite billions of online images have been embedded with visible watermarks for ownership declaration by watermarking techniques, they always suffer from a security flaw that watermarks may be affected and damaged by various watermark processing methods.}
To evaluate and improve the robustness of watermarks, a number of scientists \wh{\cite{santoyo2017automatic,pei2006novel,huang2004attacking,xu2017automatic,dekel2017effectiveness,qin2018visible,cheng2018large}} attempt to attack it by removing watermarks from images.

\begin{figure}[t]
	\begin{center}
		\subfigure[Watermarked images]{
			\includegraphics[width=0.315\textwidth]{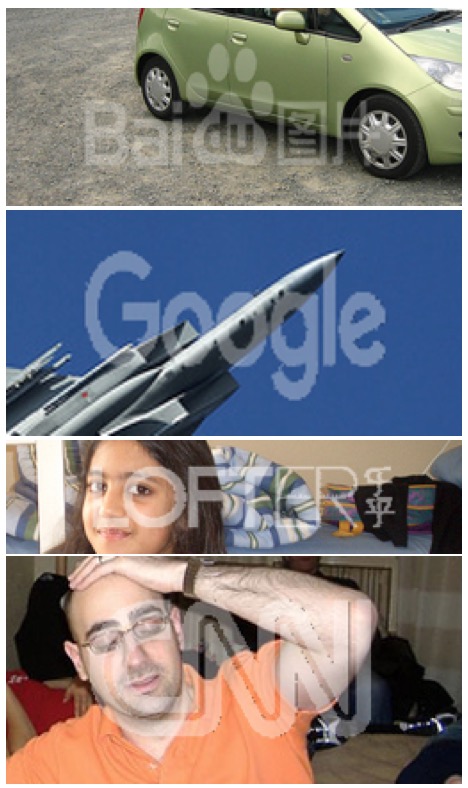}
		}
		\subfigure[Cheng et al.'s results ]{
			\includegraphics[width=0.315\textwidth]{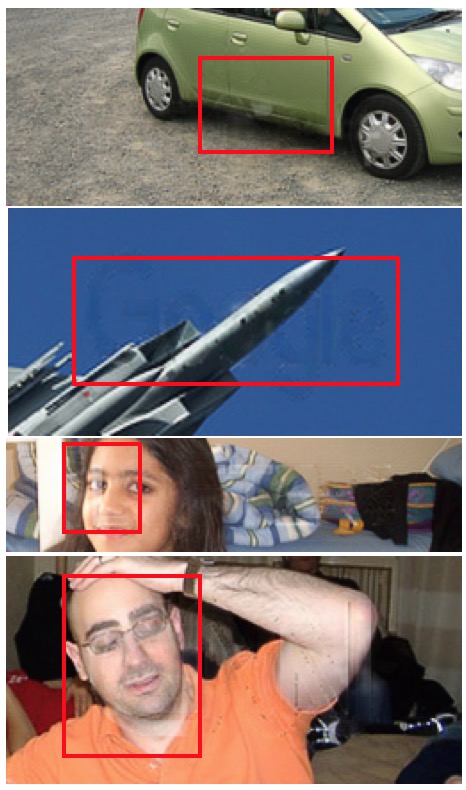}
		}
		\subfigure[Our results]{
			\includegraphics[width=0.315\textwidth]{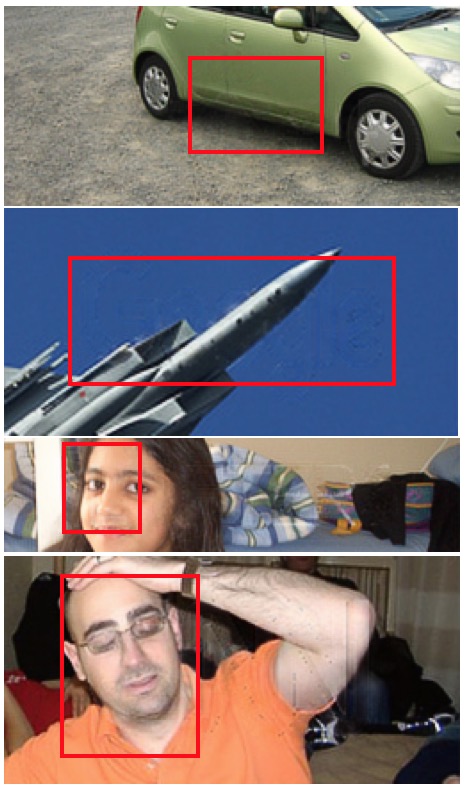}
		}
	\end{center}
	\vspace{-0.4cm}
	\caption{In real-world scenarios, visible watermarks \wh{usually} contain complex structures (a). Compared with Cheng et al.'s~\cite{cheng2018large} recovered results which remain a few residual watermark traces and are perceptually not photo-realistic (b), our framework is able to generate more convincing results (c).}
	\label{intro} \vspace{-0.3cm}
\end{figure}

Due to the \wh{existence of diverse categories and patterns} of visible watermarks, developing an advanced visible watermark removal method remains as a difficult task. More specifically, visible watermarks often contain complex structures \wh{(e.g., the texts, symbols, graphic, thin lines and shadows are diverse (Fig.~\ref{intro}(a)))}, leading to the challenge of removing unknown and diverse patterns of watermarks from images without user supervision or prior information in practical situation. 

Most of \whh{existing watermark removal methods} are \wh{unable} to tackle \whh{those aforementioned} challenges \whh{on} \wh{removing watermarks from} images. Although these models are designed for estimating and wiping off the watermark regions, they either highly depend on the prior knowledge~\cite{santoyo2017automatic,huang2004attacking,pei2006novel} or assume \wh{that} the watermarked images have the same watermark pattern~\cite{dekel2017effectiveness,xu2017automatic}, which are not suitable for removing watermarks in real-world scenarios where the watermarks may be unknown and the watermarks in different images \wh{are more likely to be}  distinct. Recently, Cheng  et al.~\cite{cheng2018large} cast the watermark removal \wh{as} an image-to-image translation problem and used a \wh{fully} convolutional architecture to transfer the watermarked pixels \wh{to} the original unmarked pixels, which provided a reasonable solution for watermark removal. \wh{However, directly training a generator with pixel-wise loss to estimate pixel relation mapping is difficult. In addition, the watermark-free images recovered by this kind of approach mostly contains a few residual watermark traces and are perceptually not photo-realistic in human visual sense (Fig.~\ref{intro}(b)).}


To make the results of watermark removal \whh{more photo-realistic} and convincing \wh{(e.g., to recover the watermarked patch without any residual watermarks and make it more photo-realistic),} in this work, we propose a new watermark removal framework with conditional generative adversarial networks (cGANs)~\cite{mirza2014conditional}. Specifically, an effective cGAN model is \whh{widely} \wh{adopted to form a framework} for photo-realistic watermark removal. To achieve this, we introduce a new loss function, \wh{consisting of} an adversarial loss and a pixel-wise content loss. \wh{In particular,} the adversarial loss \whh{working with a patch-based discriminator network} \wh{enables our method to reconstruct a photo-realistic watermark-free image}.
\whh{Here, the }\wh{ patch-based discriminator network is} conditioned on the input watermarked images \wh{and} is trained to differentiate \wh{the difference} between the recovered images and original watermark-free images. 
\wh{Additionally}, we use a content loss motivated by perceptual similarity and pixel similarity~\cite{johnson2016perceptual,cheng2018large}, which \wh{consists of the} L1 loss and \wh{the} perceptual loss. \wh{With both the adversarial loss and the content loss,} our framework is able to generate more convincing recovered results from images \wh{marked by diverse unknown watermarks} (Fig.~\ref{intro}(c)).

\wh{In summary,} our contributions are twofold. \wh{Firstly}, to the best of our knowledge, this is the \textit{first} work to exploit the concept of cGAN to design an effective framework to solve the visible watermark removal problem in \wh{a realistic setting}. Our cGAN-based watermark removal framework is much more principled than existing approaches. \wh{Secondly}, we introduce an effective watermark removal cGAN model \wh{with} a new loss function, \wh{which is comprised of} an adversarial loss and a pixel-wise content loss. \wh{This} can drive the reconstruction of \wh{the} watermark regions \wh{to be more photo-realistic}. Moreover, extensive experiments are conducted \wh{on a large-scale visible watermark dataset} for evaluation. \wh{The} results demonstrate that \wh{our} proposed model is capable of addressing the visible watermark removal problem confronted in real-world scenarios, achieving more convincing reconstruction \wh{than} state-of-the-art methods.

\section{Methodology}
\wh{In this section, we present our watermark removal framework which is build based on the concept of cGANs \cite{mirza2014conditional,isola2017image}.}
\wh{In recent, the cGANs~\cite{isola2017image} are commonly adopted to reconstruct the hidden information which is obscured in original image.} In this work, as we aim at restoring the original images \wh{from} the watermarked images, we adopt the idea of the cGANs and propose a cGANs-based framework \wh{for watermark removal}. The architectures of our proposed watermark removal cGANs is \whh{illustrated in} Fig.~\ref{framework}.

Our network \wh{mainly} embodies a generator and a discriminator. \wh{In the generator,} we leverage a U-net based architecture~\cite{ronneberger2015u} to transform a watermarked image \wh{to} a watermark-free one. \wh{In our discriminator, we use a patch-based classifier~\cite{isola2017image} conditioned on the input watermarked images to distinguish those recovered images generated by the generator from the ground-truth watermark-free images in a patch level.}

\wh{More specifically, our network takes as input a watermarked image and exploits the generator $\mathbf{G}$ to generate a photo-realistic watermark-free image. To enable the image \whh{restored} by the generator to be similar to the ground-truth watermark-free image as much as possible, we introduce \whh{a new objective which is} the combination of the L1 loss, perceptual loss and the patch-based adversarial loss to restrain the training of the generator. In the meanwhile, an adversarially trained discriminator $\mathbf{D}$ is employed to detect the ``fake'' images (i.e., the images which is generated by the generator and is not distinguished as the real watermark-free images) from those ground-truth images (i.e., the real watermark-free images). We detail the adversarial network architecture and loss functions individually in Section 2.1 and 2.2.}



\subsection{Adversarial Network Architectures}
We \wh{formulate} our generator and discriminator architecture inspired by~\cite{isola2017image}. Fig.~\ref{framework} shows the details of our architectures, and the key features will be illustrated below.

\begin{figure}[t]\vspace{0cm}
	\begin{center}
		{
			\includegraphics[width=0.95\textwidth]{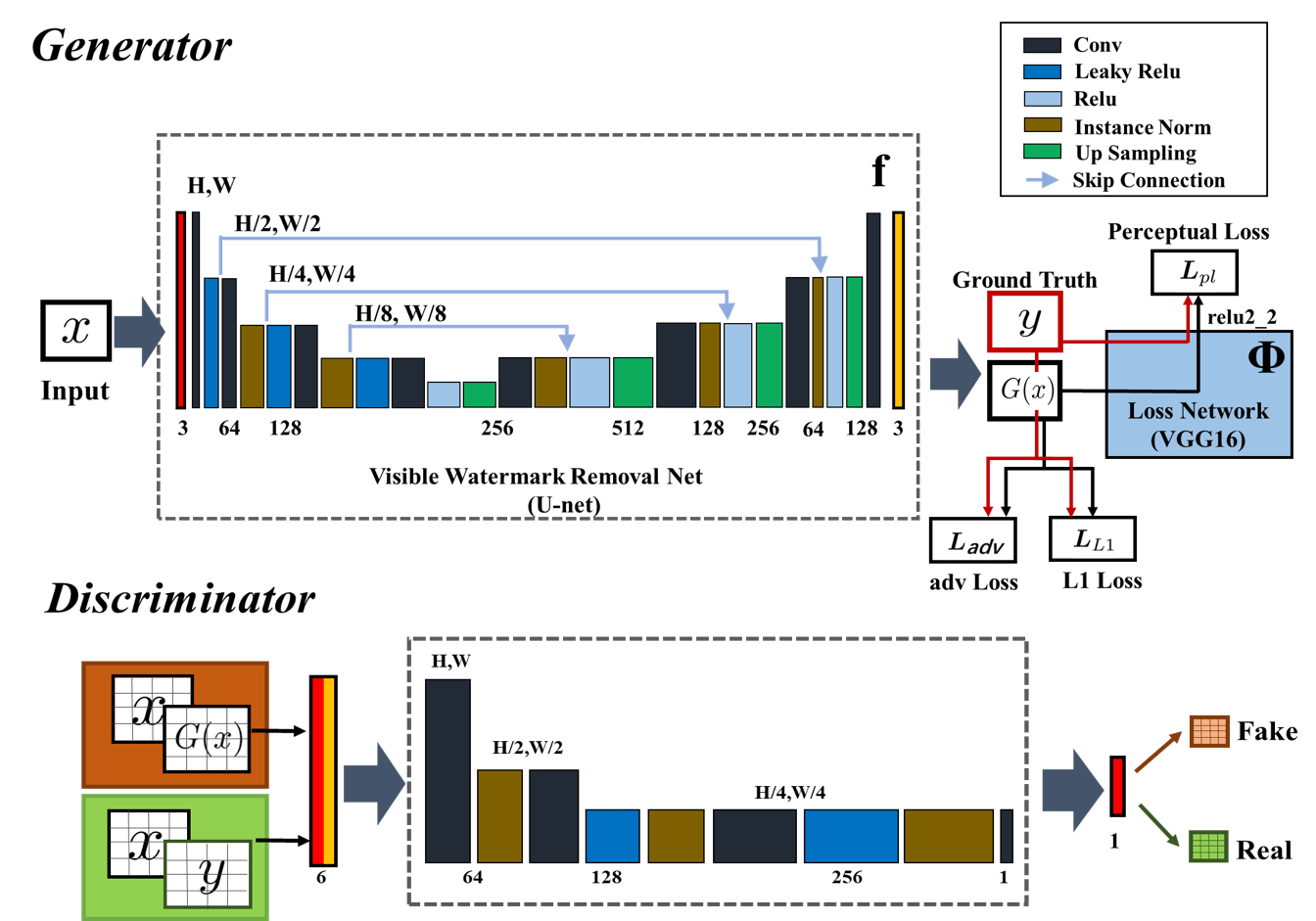}
		}
	\end{center}
	\vspace{-0.1cm}
	\caption{The architecture of our visible watermark removal framework.}
	\label{framework} \vspace{-0.2cm}
\end{figure}

\noindent\textbf{Generator (U-net).} \wh{Typically, watermark removal can be cast as an image-to-image translation problem~\cite{cheng2018large}. Analyzing \wh{the} watermark removal task, we \wh{find that} the unmarked image areas share the pixel values as the input while the watermarked pattern needs to be removed to meet the visual requirements.} Unlike \wh{the} general encode-decode structure~\cite{badrinarayanan2017segnet}, which directly \wh{transforms} an image in source domain to a target image through a series of convolution module in the network, we adopt \wh{a} U-net based architecture as our generator in our work, \wh{followed by} the \wh{fully} convolutional network proposed in~\cite{cheng2018large}. \wh{In our system, the} U-net takes \wh{the advantage} of its skip connection structure, which combines the low level feature \wh{and the} high level features, \wh{allowing} the sharing of global information and edge details between the input and the output. Specifically, our generator comprises \wh{of} six standard modules, \whh{which are down-blocks or up-blocks.} In down-blocks, the channels of feature map \wh{are doubled} and the its side size \wh{is reduced} by half, \wh{while} the up-blocks go the opposite. \wh{In addition}, there are skip connections between \wh{every} $i^{th}$ layer and the ${(n-i)}^{th}$ layer, where n is the total number of layers. \wh{Each} skip connection simply concatenates all channels \wh{of} the $i^{th}$ layer with those of the ${(n-i)}^{th}$ layer as \wh{shown} in Fig.~\ref{framework}.

\noindent\textbf{Discriminator (patch-based).} \cdn{Different from \whh{common GAN discriminators}, which \whh{map} the input into \whh{one scalar representing the probability of the input sample attributed to ``real''}. In our work, \whh{as shown} in Fig.~\ref{framework}, we employ \whh{the} patch-based network~\cite{isola2017image} as our discriminator. The structure of our discriminator is a full convolutional network, which \whh{maps} the combination of watermarked image and watermark removed image to a feature map, \whh{representing} the \whh{class} probabilities \whh{i.e., ``fake'' or ``real''} of the patches of the input. Since the point in the feature map can be traced back to the receptive field in the original image, thus \whh{each} value in the output matrix refers the probability that the patch in the original image is ``real'', and we calculate the probability of the input image is ``real'' as the average of the all patches are ``real''.}

\cdn{Observing the watermark images, we can find that the difference between the watermarked image and the original watermark-free one solely exists in some parts of the image. Since the watermarked area is relatively small compared with the whole image, it is critical for the discriminator to identify the most different patches of two input images and focus more on minimizing loss of these patches. Thus, introducing patch-based discriminator can make our cGANs based network be more powerful for removing visible watermark.}


\subsection{Objective Function}

\wh{The objective functions play a very important role in training the network. In this work, we aim to learn a solution $\mathbf{G}^{*}$ to minimize the loss function defined as below:
\begin{equation}\label{eq:objective}
\boldsymbol{G}^{*}=\arg\underset{G}{\min}\,\underset{D}{\max}\ \underset{adversarial\ loss}{\underbrace{\boldsymbol{L}_{adv}(G,D)}}+\underset{content\ loss}{\underbrace{\alpha \boldsymbol{L}_{l_{1}}(G)+\beta \boldsymbol{L}_{per}(G)}}.
\end{equation}
During each stage of training our cGANs-based watermark removal model the generator $\mathbf{G}$ and the discriminator $\mathbf{D}$ are trained alternately, where $\mathbf{G}$ is trained to minimize this objective against an adversarial discriminator $\mathbf{D}$, which is trained to maximize the loss.}


\wh{Specially, the generator $G$ is trained by minimizing the loss. The task of generator is not only to fool the discriminator but also \wh{to generate a image that is closed to the ground truth watermark-free image in visual}. Therefore, the objective of the generators consists of a content loss and an adversarial loss, where the perceptual loss and \wh{$l_1$} loss comprise the content loss. In the Eq. (\ref{eq:objective}), $\boldsymbol{L}_{l_1}$ ,$\boldsymbol{L}_{per}$ and $\boldsymbol{L}_{adv}$ refer to the $l_1$ loss, perceptual loss and adversarial loss respectively, where $\alpha$ \wh{and} $\beta$ are weights to \wh{balance} the \wh{$l_1$} loss, perceptual loss and adversarial loss. 
The discriminator $\mathbf{D}$ is trained alternately to avoid being fooled by the generators by distinguishing the inputs as either real or fake. Thus, the adversarial loss is defined as the opposite loss function as \wh{that} in \wh{the} generator. We detail the content loss and adversarial loss in the following sections.}

\noindent\textbf{Content loss.}
At present\wh{,} the most commonly used content loss \wh{in} image-to-image tasks is MSE loss. \wh{It obtained the} state-of-art PSNR results \wh{in} many image-to-image task such as super-resolution and image style translation and etc. However, it \wh{is proved} to be blur occurrence \wh{in} the \wh{generated} images and \wh{the} output results \wh{do not satisfy human visual sense}. To solve this problem, we use the $l_1$ distance rather than the $l_2$, which is defined as:
\begin{equation}
\boldsymbol{L}_{l_1}(G)=\left \| G(x)-y \right \|_{1},
\end{equation}
$G(x)$ denotes the output of generators and $y$ denote the ground truth watermark-free image. \wh{Apart from the $l_1$} loss, it is beneficial to \wh{inject the} perceptual loss for watermark removal~\cite{cheng2018large}. The perceptual loss function of our network can be expressed as:
\begin{equation}
\boldsymbol{L}_{per}^{\Phi,j}(G)=\frac{1}{C_{j}H_{j}W_{j}}\left \| \Phi _{j}(G(x))-\Phi _{j}(y) \right \|_{2}^{2}.
\end{equation}
\wh{Here, we} define $\Phi$ as the convolutional transformation for calculating the perceptual loss, which refers to \wh{the} pertrained $VGG16$~\cite{simonyan2014very} in our work. The feature size of the $j_{th}$ convolutional layer of loss network is $C_{j}\times H_{j}\times W_{j}$. \wh{Specifically, the} weight of the $VGG16$ is \wh{frozen} and \wh{the outputs of the $\mathbf{relu2\_2}$ are extracted as features} to calculate the semantic difference between the input and output of generator \wh{as}~\cite{johnson2016perceptual}.

\noindent\textbf{Adversarial loss.} Different from \wh{the} general GAN~\cite{goodfellow2014generative}, the discriminator of conditional GAN \wh{observes not only} the output of $G$, but also the input $x$. Mathematically, the adversarial loss can be formulated as:
\begin{equation}
\boldsymbol{L}_{adv}(G,D)=\mathbb{E}_{x,y}[\log D(x,y)]+\mathbb{E}_{x}[\log(1-D(x,G(x)))]\wh{.}
\end{equation}
Here, rather than training $G$ to minimize $\log(1-D(x,G(x)))$, we instead train to maximize $\log(D(x,G(x)))$~\cite{goodfellow2014generative}.

\section{Experiments}

\subsection{Datasets and Settings}

\begin{figure}[t]\vspace{-0cm}
	\begin{center}
		{
			\includegraphics[width=1\textwidth,height=0.48\textheight]{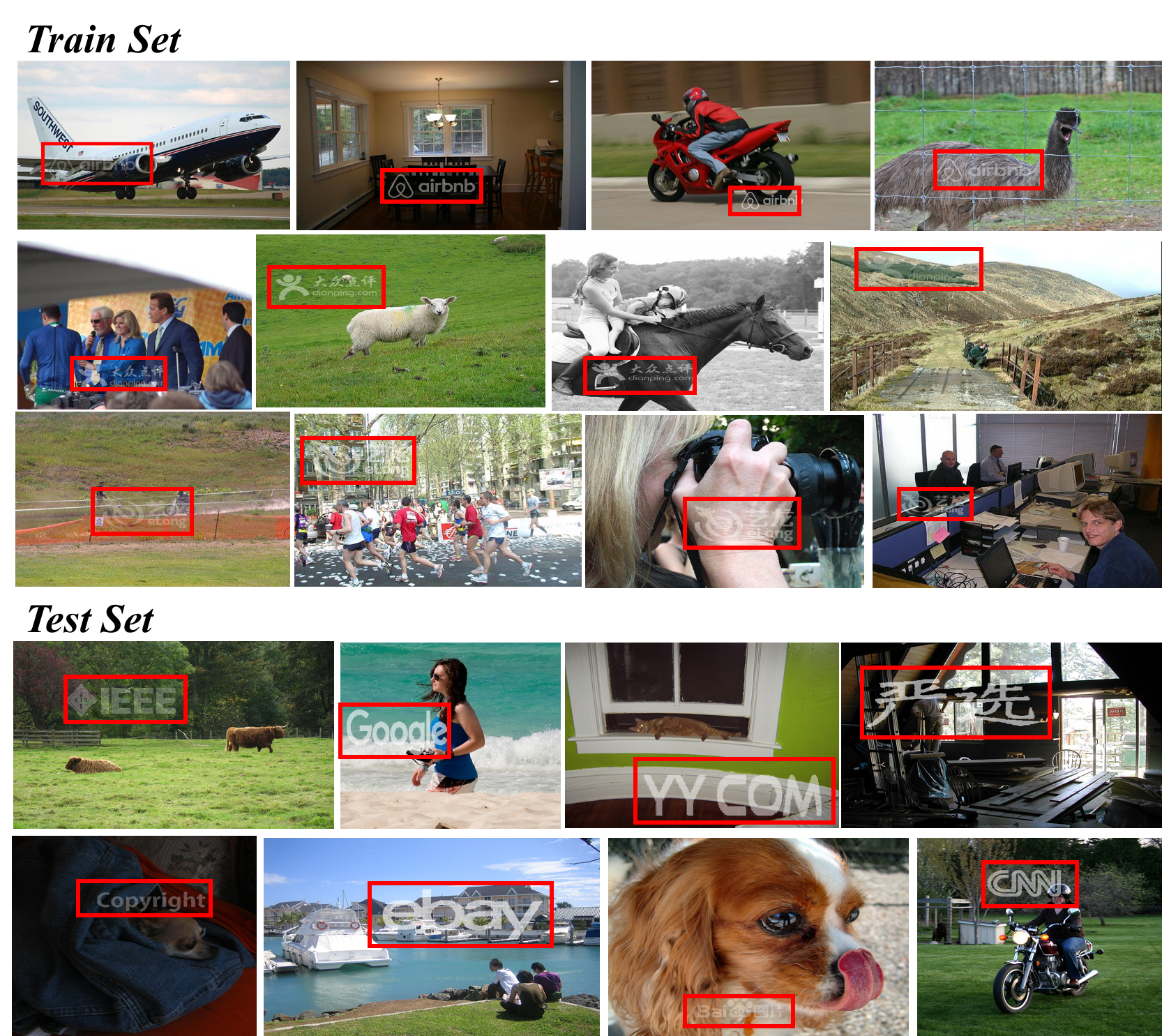}
		}
	\end{center}
	\vspace{-0.1cm}
	\caption{The example images of LVW dataset.}
	\label{dataset} \vspace{-0.2cm}
\end{figure}

\noindent\textbf{Dataset.} To evaluate the performance of our framework on the large visible watermark image dataset, we \wh{conduct extensive experiments on} the Large-scale Visible Watermark Dataset (LVW)~\cite{cheng2018large}, containing 60k watermarked images made of 80 watermarks, with 750 images per watermark.
\wh{In this dataset, the} original images in the training and \wh{testing} sets are randomly chosen from the train/val and test sets in PASCAL VOC2012 dataset with replacement\wh{,} respectively. \wh{The 80 categories of watermarks \wh{covering} a vast quantity of patterns (e.g., the watermarks contain English and Chinese), are collected from renowned E-commercial brand, websites, organization, personal, and etc.} Moreover, the size, location and transparency of each watermark in different images are distinct and set randomly. Example images of LVW dataset are shown in Fig.~\ref{dataset}.

\noindent\textbf{Training details.}
PyTorch platform was applied to construct the proposed deep architecture. All experiments are conducted on a computer cluster equipped with NVIDIA Tesla K80 GPU with 12GB memory.  To optimize the generative adversarial networks, we follow \wh{the training strategy in}~\cite{goodfellow2014generative} to alternate between one gradient descent step on the discriminator, then one step on generator. \wh{During training, we use mini batch SGD (the batch size is set to be 1) and apply the Adam solver~\cite{kingma2014adam}, with \wh{the initial learning rate of} 2e-4 and momentum parameters (i.e., $\beta1$ = 0.5, $\beta2$ = 0.999)}. And we evaluated the proposed framework with default setting ($\alpha$ = 10, $\beta$ = 1e-4).

\noindent\textbf{\wh{Evaluation setting and metrics}.}
In our experiments, the watermarks in training set are different from those in \wh{testing} set. In LVW dataset, \wh{around} $80\%$ sorts of watermark are used for training, and the remaining $20\%$ \wh{are} for test (see Fig.~\ref{dataset}), \wh{which is the same as ~\cite{cheng2018large}} . This setting meets the \wh{requirements} of unknown watermarks removal in real-world scenarios well. \wh{Both of the Peak signal to noise ratio (PSNR) and structural dissimilarity image index (DSSIM), measuring the similarity between the recovered image and the ground truth one, are adopted as evaluation metrics by previous work (e.g.,~\cite{dekel2017effectiveness,cheng2018large}).} However, \wh{both} metrics fail to capture and accurately assess image quality with respect to the human visual system~\cite{ledig2017photo}. \wh{Therefore, in addition to two aforementioned evaluation metrics, we used \wh{the} mean opinion score (MOS) testing~\cite{ledig2017photo} to further quantify the ability of different methods of reconstructing photo-realistic and convincing watermark-free images from watermarked images.} Specifically, we asked 10 raters to assign an integral score from 1 (bad quality, i.e.\wh{,} watermarked images) to 5 (excellent quality, i.e.\wh{,} original images) to the recovered images for assessing the quality of the recovered images.

\subsection{Results and Analysis}

\noindent\textbf{Analysis of the objective function.}
The objective function of our framework has three components terms in Eq. \wh{(\ref{eq:objective})}, including the \wh{$l_1$} loss term, the perceptual loss term and the adversarial loss term.
In this section, we \wh{conduct} experiments to analyze the effect of \wh{these} loss terms. 

\begin{table}
	\centering
	\vspace{-0cm}
	\caption{Evaluation of different loss functions.}
	\setlength{\tabcolsep}{15pt}
	\begin{tabular}{c|c|c|c}
		\hline \hline
		Loss   &PSNR &DSSIM  & MOS \\\hline \hline
		L1  & 30.42 & 0.045 & 2.85 \\
		Perceptual  & 29.86 & 0.051 & 3.17 \\
		L1 + Perceptual   & \textbf{30.86} & \textbf{0.043} & 3.23 \\
		L1 + Perceptual + GAN  &30.33 &0.049 &3.31 \\
		L1 + Perceptual + cGAN  &30.69 &0.045 &\textbf{4.08}\\
		\hline
	\end{tabular}\label{Evaluation_loss}\vspace{-0cm}
\end{table}

In Table~\ref{Evaluation_loss}, `L1' and `Perceptual' indicate that the generator network \wh{only using the $l_1$ loss (Eq. (2)) and perceptual loss (Eq. (3)), respectively.}
 `L1 + Perceptual' \wh{represents} the generator network using the \wh{combination loss of} the \wh{$l_1$} loss as well as \wh{the} perceptual loss. As shown in Table~\ref{Evaluation_loss},  `L1 + Perceptual' performs clearly better than the `L1' and `Perceptual', demonstrating that the combination of \wh{$l_1$} loss and perceptual loss can incorporate the strength of both losses to reconstruct the fine details. 

To evaluate the effect of adversarial loss term, we further show the performance of the combination of adversarial loss and L1 loss and the perceptual loss. Specifically, we \wh{compare the} \wh{model} using a discriminator conditioned on the input (adversarial loss of cGAN, Eq. (4)) \wh{with the model} using an unconditional discriminator (adversarial loss of GAN). \wh{They are} respectively denoted as `L1 + Perceptual + cGAN' and `L1 + Perceptual + GAN' in Table~\ref{Evaluation_loss}. Although \wh{combining the $l_1$ loss and the perceptual loss} with the adversarial loss \wh{causes a slight drop in} the PSNR and DSSIM values, \wh{it} achieves higher MOS scores, indicating that the recovered results are more photo-realistic. \wh{The reason is} that \wh{the} GAN-based procedure encourages the reconstructions to move towards regions with high probability of containing photo-realistic images in searching space and thus closer to the convincing results. Moreover, the results in Table~\ref{Evaluation_loss} also show clearly that cGAN performs much better than GAN, \wh{verifying} the effectiveness of \wh{the conditional discriminator}. This suggests that it is important that the loss measure the quality of the match between input (watermarked images) and output recovered images.

\noindent\textbf{Evaluation of the patch-based discriminator.} \wh{As the patch-based discriminator is essential in our proposed framework to model the discriminative information \wh{of} local image patches}. To \wh{investigate} the \wh{effect} of patch-based discriminator, we compared \wh{our model with the model using the} conventional image-based discriminator, which \wh{classifies} if whole image region is real or fake (i.e., in image level). Note that\wh{,} in this section all experiments \wh{are conducted with} the \wh{`L1 + perceptual + cGAN'} loss (Eq. \wh{(\ref{eq:objective})}).

\begin{table}\vspace{-0cm}
	\centering
	\caption{Evaluation of different discriminators.}
	\setlength{\tabcolsep}{15pt}
	\begin{tabular}{c|c|c|c}\hline \hline
		Discriminator   &PSNR &DSSIM  & MOS \\\hline \hline
		 image-based &29.72  &0.052  &3.46  \\
		 patch-based  & \textbf{30.69} & \textbf{0.045} & \textbf{4.08} \\
		 \hline
	\end{tabular}
	\label{Evaluation_discriminator}\vspace{-0cm}
\end{table}

The results are shown in Table~\ref{Evaluation_discriminator}. Compared with our patch-based discriminator, the image-based discriminator gets a considerably \wh{worse} performance. 
Specifically, the image-based discriminator identify the difference between two images in a image level, which alleviate the effect of the difference in local areas and hamper the results (i.e., the generated images are not photo-realistic). In addition, the image-based discriminator has much more parameters and deeper than the patch-based discrmininator. This can slow down the speed of the watermark removal model and it is harder to train the model, which makes this kind of discriminator unscalable to the real-world application. In other words, as the patch-based discriminator is a light weight model, it can run faster even on arbitrarily large image and it is shown to perform better. This strongly suggests that our proposed model is more suitable to be applied to train the cGANs-based model for visible watermarks removal in realistic data.

\noindent\textbf{Comparison with state-of-the-art.} To justify the effectiveness of the proposed model, we performed experiments \wh{to compare our method with Cheng et al.~\cite{cheng2018large}}. As shown in Table~\ref{comparison_sota}, our model obtained the comparable results in PSNR and DSSIM, and the MOS results \wh{indicate} that our method outperforms Cheng et al. by a large margin in human visual system. 

\begin{table}\vspace{-0cm}
	\centering
	\caption{Comparisons with state-of-the-art method on LVW dataset.}
	\setlength{\tabcolsep}{15pt}
	\begin{tabular}{c|c|c|c|c}\hline \hline
		Metrics               &Input    &Cheng \textit{et al}.~\cite{cheng2018large}       &Ours  & Ground truth \\\hline\hline
		PSNR                 &20.65      &30.86          &30.69  &$\infty$\\
		DSSIM                 &0.103     &0.043          &0.045  &0\\
		MOS                 &1.0        &3.23          &\textbf{4.08}  &5.0\\
		\hline
	\end{tabular}
	\label{comparison_sota}\vspace{-0cm}
\end{table}

\begin{figure}[t]
	\begin{center}
		{
			\includegraphics[width=1\textwidth,height=0.5\textwidth]{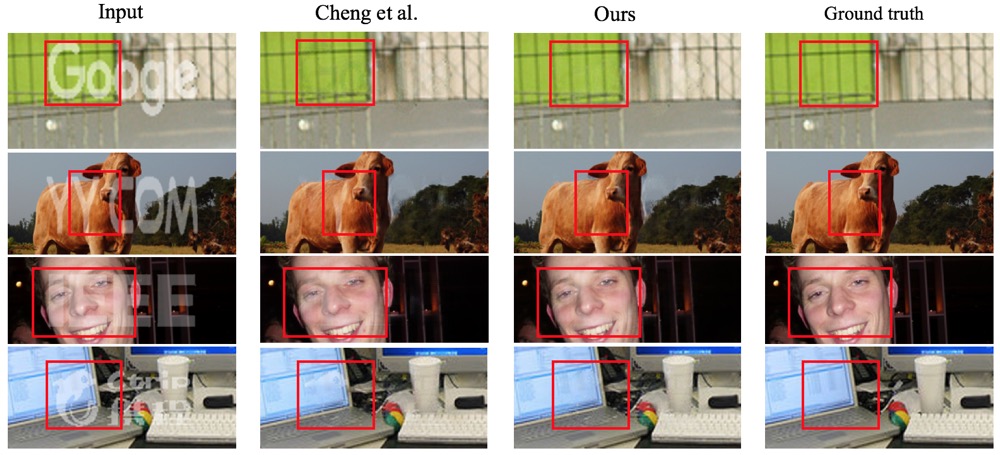}
		}
	\end{center}
	\vspace{-0cm}
	\caption{Example results on watermark removal. From left to right: watermarked images, Cheng et al.'s results, our results, ground truth. This experiment demonstrates that our method is effective to generate more photo-realistic recovered images. (best view in color)}
	\label{exper_results} \vspace{0cm}
\end{figure}

We also visualized the watermark removal results of test examples in LVW dataset and show them in Fig.~\ref{exper_results}. The results in the figure further demonstrate that the performance of the proposed method is noticeably convincing than the \wh{ones} of existing methods, suggesting that the adversarial model is more suitable for solving the visible watermark removal problem.

\subsection{Discussion and Future Work}
Our experiments show that our proposed framework can effectively remove the unknown and diverse visible watermarks, resulting \wh{more satisfactory recovered images}. The focus of this work was the photo-realistic quality of reconstruction rather than \wh{obtaining better} performance \wh{in} standard quantitative \wh{evaluation metrics} such as PSNR and SSIM, which \wh{can} not accurately capture and evaluate the quality of images associated with the human visual system. The \wh{experimental results} further \wh{verify} that a deep convolutional architecture using the concept of cGANs \wh{to form an} adversarial loss is useful for photo-realistic watermark removal in real-world scenarios. Significantly, our original intention is to increase \wh{the} awareness \wh{on the copyrights of} online images, reminding that visible \wh{watermarks} should be designed to be more resistant \wh{against} removal attacking. Developing a more robust watermarking technique for copyright protection is challenging and part of future work.

\section{Conclusion}

In this work, we introduced a new watermark processing framework for more photo-realistic visible watermark removal, which augments the conventional L1 and perceptual loss function with an adversarial loss by training a conditional generative adversarial network. 
The proposed model is able to drive the reconstruction of watermark regions towards the photo-realistic results producing perceptually more convincing solutions. Extensive experiments are conducted on a large-scale visible watermark dataset to verify the feasible of our method. Experimental results clearly demonstrated the superior performance of the proposed framework compared to existing methods. 

\section*{Acknowledgment}
This work was supported by NSFC(U1811461). Xiang Li and Chan Lu equally contributed to this work.

{
	\bibliographystyle{splncs03}
	\bibliography{egbib}

\begin{thebibliography}{10}
\providecommand{\url}[1]{\texttt{#1}}
\providecommand{\urlprefix}{URL }

\bibitem{badrinarayanan2017segnet}
Badrinarayanan, V., Kendall, A., Cipolla, R.: Segnet: A deep convolutional
  encoder-decoder architecture for image segmentation. IEEE transactions on
  pattern analysis and machine intelligence  39(12),  2481--2495 (2017)

\bibitem{cheng2018large}
Cheng, D., Li, X., Li, W.H., Lu, C., Li, F., Zhao, H., Zheng, W.S.: Large-scale
  visible watermark detection and removal with deep convolutional networks. In:
  Chinese Conference on Pattern Recognition and Computer Vision (PRCV). pp.
  27--40. Springer (2018)

\bibitem{dekel2017effectiveness}
Dekel, T., Rubinstein, M., Liu, C., Freeman, W.T.: On the effectiveness of
  visible watermarks. In: Proceedings of the IEEE Conference on Computer Vision
  and Pattern Recognition. pp. 2146--2154 (2017)

\bibitem{goodfellow2014generative}
Goodfellow, I., Pouget-Abadie, J., Mirza, M., Xu, B., Warde-Farley, D., Ozair,
  S., Courville, A., Bengio, Y.: Generative adversarial nets. In: Advances in
  neural information processing systems. pp. 2672--2680 (2014)

\bibitem{huang2004attacking}
Huang, C.H., Wu, J.L.: Attacking visible watermarking schemes. IEEE
  transactions on multimedia  6(1),  16--30 (2004)

\bibitem{isola2017image}
Isola, P., Zhu, J.Y., Zhou, T., Efros, A.A.: Image-to-image translation with
  conditional adversarial networks. In: Proceedings of the IEEE conference on
  computer vision and pattern recognition. pp. 1125--1134 (2017)

\bibitem{johnson2016perceptual}
Johnson, J., Alahi, A., Fei-Fei, L.: Perceptual losses for real-time style
  transfer and super-resolution. In: European conference on computer vision.
  pp. 694--711. Springer (2016)

\bibitem{kingma2014adam}
Kingma, D.P., Ba, J.: Adam: A method for stochastic optimization. arXiv
  preprint arXiv:1412.6980  (2014)

\bibitem{ledig2017photo}
Ledig, C., Theis, L., Husz{\'a}r, F., Caballero, J., Cunningham, A., Acosta,
  A., Aitken, A., Tejani, A., Totz, J., Wang, Z., et~al.: Photo-realistic
  single image super-resolution using a generative adversarial network. In:
  Proceedings of the IEEE conference on computer vision and pattern
  recognition. pp. 4681--4690 (2017)

\bibitem{mirza2014conditional}
Mirza, M., Osindero, S.: Conditional generative adversarial nets. arXiv
  preprint arXiv:1411.1784  (2014)

\bibitem{pei2006novel}
Pei, S.C., Zeng, Y.C.: A novel image recovery algorithm for visible watermarked
  images. IEEE Transactions on Information Forensics and Security  1(4),
  543--550 (2006)

\bibitem{qin2018visible}
Qin, C., He, Z., Yao, H., Cao, F., Gao, L.: Visible watermark removal scheme
  based on reversible data hiding and image inpainting. Signal Processing:
  Image Communication  60,  160--172 (2018)

\bibitem{ronneberger2015u}
Ronneberger, O., Fischer, P., Brox, T.: U-net: Convolutional networks for
  biomedical image segmentation. In: International Conference on Medical image
  computing and computer-assisted intervention. pp. 234--241. Springer (2015)

\bibitem{santoyo2017automatic}
Santoyo-Garcia, H., Fragoso-Navarro, E., Reyes-Reyes, R., Sanchez-Perez, G.,
  Nakano-Miyatake, M., Perez-Meana, H.: An automatic visible watermark
  detection method using total variation. In: 2017 5th International Workshop
  on Biometrics and Forensics (IWBF). pp. 1--5. IEEE (2017)

\bibitem{simonyan2014very}
Simonyan, K., Zisserman, A.: Very deep convolutional networks for large-scale
  image recognition. arXiv preprint arXiv:1409.1556  (2014)

\bibitem{xu2017automatic}
Xu, C., Lu, Y., Zhou, Y.: An automatic visible watermark removal technique
  using image inpainting algorithms. In: 2017 4th International Conference on
  Systems and Informatics (ICSAI). pp. 1152--1157. IEEE (2017)

\end{thebibliography}
}

\end{document}